\documentclass[conference]{IEEEtran}
\IEEEoverridecommandlockouts
\usepackage{cite}
\usepackage{amsmath,amssymb,amsfonts}
\usepackage{algorithm}
\usepackage{algorithmic}
\usepackage{graphicx}
\usepackage{subcaption}
\usepackage{textcomp}
\usepackage{xcolor}
\usepackage{enumerate}
\usepackage{booktabs}
\usepackage{multirow}

\def\BibTeX{{\rm B\kern-.05em{\sc i\kern-.025em b}\kern-.08em
    T\kern-.1667em\lower.7ex\hbox{E}\kern-.125emX}}
\begin{document}

\title{BiFedKD: Bidirectional Federated Knowledge Distillation Framework for Non-IID and Long-Tailed ECG Monitoring\\

\thanks{
This work is supported by the Nanyang Assistant Professorship and the Schmidt AI in Science Postdoctoral Fellowship.

Zixuan Shu, Tiancheng Cao, and Hen-Wei Huang are with the School of Electrical and Electronic Engineering, Nanyang Technological University, Republic of Singapore. Hen-Wei Huang is also with the Lee Kong Chian School of Medicine, Nanyang Technological University, Republic of Singapore (e-mail: SHUZ0004@e.ntu.edu.sg; tiancheng.cao@ntu.edu.sg; henwei.huang@ntu.edu.sg).
}
}

\author{
\IEEEauthorblockN{
Zixuan Shu,
Tiancheng Cao,~\IEEEmembership{Member,~IEEE,}
and Hen-Wei Huang,~\IEEEmembership{Member,~IEEE}
}
}

\maketitle


\begin{abstract}
Electrocardiogram (ECG) monitoring in Internet of Medical Things (IoMT) networks is constrained by strict data-sharing regulations and privacy concerns. 
Federated learning (FL) enables collaborative learning by keeping raw ECG data on devices, but frequent transmissions of high-dimensional model updates incur heavy per-round traffic over bandwidth-limited links. 
To alleviate this bottleneck, federated distillation (FD) replaces parameter exchange with logit-based knowledge transfer. 
However, the performance of FD often degrades under the non-independent and identically distributed (non-IID) and long-tailed label distributions in ECG deployments. 
To address these challenges, we propose a bidirectional federated knowledge distillation (BiFedKD) framework that employs an aggregation-by-distillation pipeline with temperature scaling to produce a stable global distillation signal for cross-client alignment. 
Experiments on the MIT-BIH Arrhythmia dataset show that BiFedKD improves accuracy and Macro-F1 over the baseline by $3.52\%$ and $9.93\%$, respectively. Moreover, to reach the same Macro-F1, BiFedKD reduces communication overhead by $40\%$ and computation cost by $71.7\%$ compared with the baseline.
\end{abstract}

\begin{IEEEkeywords}
ECG Monitoring, Federated Knowledge Distillation, Internet of Medical Things
\end{IEEEkeywords}

\section{Introduction}
In Internet of medical things (IoMT) networks, electrocardiogram (ECG) data are inherently distributed across multiple resource constrained edge devices \cite{neucom.2023.126719}. 
Meanwhile, heightened privacy concerns and regulations such as the General Data Protection Regulation (GDPR) and the Health Insurance Portability and Accountability Act (HIPAA) impose strict limitations on data sharing, which hinders centralized training of large-scale models \cite{yang2019federated}.
Federated learning (FL) has therefore emerged as a privacy-preserving paradigm by keeping raw ECG data on devices and coordinating training via server aggregation \cite{9794622}. 
However, from the networking perspective, FL critically relies on frequent uplink and downlink transmissions of high-dimensional model parameters, which can dominate per-round traffic and become prohibitive under bandwidth-limited links and heterogeneous device capabilities in IoMT \cite{ni2024fedsl}. 

To alleviate the communication bottleneck of parameter exchange, 
federated distillation (FD) has been introduced as a communication-efficient alternative method that leverages knowledge distillation (KD)
\cite{hinton2015distill} 
to enable cross-device knowledge transfer via logits \cite{jeong2018communication}. 
Compared with FL, FD avoids the explicit parameter transmission, thereby substantially reducing per-round communication cost \cite{lin2020ensemble}.

\begin{figure}[t]
	\centering
	\includegraphics[width=0.48\textwidth]{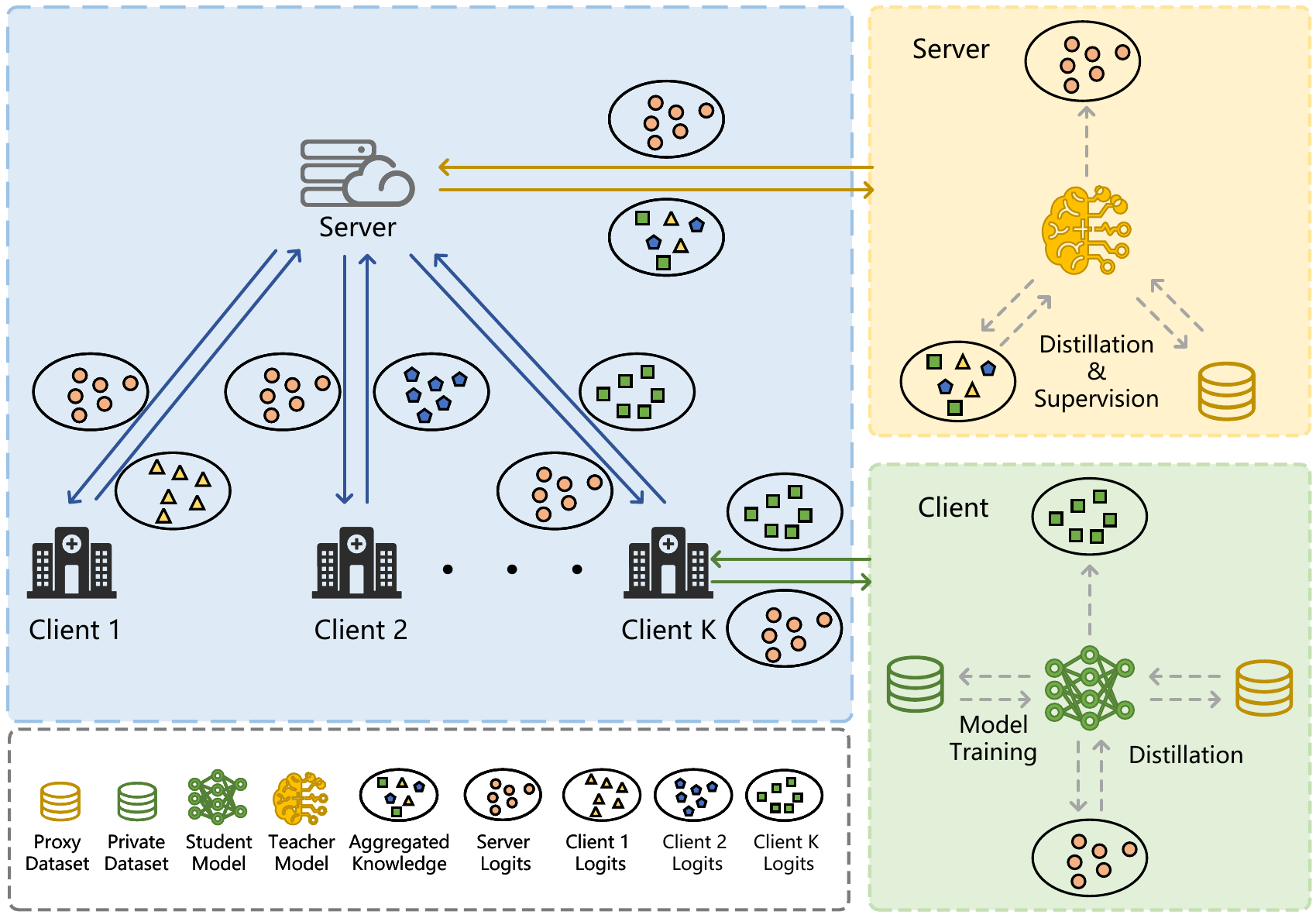}
	\caption{The framework of BiFedKD.}
	\label{fig_framework}
\end{figure}

However, the practical performance of FD in IoMT is often limited by data heterogeneity.
Due to variations in patient populations and pathological conditions, ECG data collected across IoMT devices are strongly non-IID and long-tailed \cite{Jimenez2023fedecg},
where normal sinus beats typically dominate, whereas many arrhythmia types occur sparsely.
In such settings, logits aggregation in FD can be dominated by head classes and majority clients, biasing the server-side soft targets and providing weak supervision for tail or even missing classes.

To address these challenges, we propose a bidirectional federated knowledge distillation (BiFedKD) framework for IoMT ECG monitoring that operates effectively under non-IID, long-tailed label distributions and stringent resource constraints.
In contrast to traditional FD methods such as FedMD \cite{fedmd}, which directly averages uploaded logits to construct consensus soft targets, BiFedKD adopts an aggregation-by-distillation pipeline that decouples client-side heterogeneity from the supervision signal delivered to clients.
Specifically, to mitigate over-confident aggregated predictions, the server applies temperature scaling to the aggregated logits \cite{guo2017calibration}, smoothing prediction sharpness and reducing majority class bias.
Moreover, to limit the impact of noisy or biased client outputs on aggregation, BiFedKD introduces a server-side teacher that absorbs the aggregated knowledge via distillation and produces refined global soft targets for top-down supervision, improving robustness under severe heterogeneity.

In IoMT deployments, uplink transmissions often constitute the primary bottleneck, and on-device training can be prohibitive for resource-limited edge devices \cite{wu2023efficient}. Therefore, accuracy improvements alone are insufficient without quantifying the associated communication and computation budgets. Accordingly, we evaluate the communication cost and computation overhead of our proposed framework in comparison with FedMD and FedAvg \cite{mcmahan2016communication}.
Moreover, experiments on the MIT-BIH Arrhythmia Dataset~\cite{moody2001mit} show that BiFedKD mitigates the adverse effects of long-tailed label imbalance and non-IID data heterogeneity in resource-limited IoMT settings.

Our contributions can be briefly summarized as follows.

\begin{itemize}
    \item We propose a resource-efficient FD framework to address long-tailed, non-IID ECG monitoring issues in IoMT systems. Using a server-side teacher model to provide top-down supervision to clients, improving cross-client knowledge alignment.
    \item We theoretically analyze the communication overhead and computation cost of BiFedKD and evaluate its performance under resource-constrained IoMT settings.
    \item Experiments on the MIT-BIH Arrhythmia Dataset verify that BiFedKD can effectively mitigate long-tailed class imbalance and non-IID client heterogeneity. 
\end{itemize}

\section{Bidirectional Federated Knowledge Distillation Framework}
Fig. \ref{fig_framework} illustrates the proposed BiFedKD framework.
We consider a system consisting of a server and $K$ clients, indexed by $\mathcal{K}=\{1,2,\dots,K\}$.
Each client $k\in\mathcal{K}$ maintains a student model with parameters $w_k$ and a private labeled dataset $\mathcal{D}_k$, where both raw samples and ground-truth labels remain local.
Moreover, the server and all clients share access to a public proxy dataset $\mathcal{D}_p$.
The server maintains a teacher model with parameters $w_s$, which provide global soft targets to supervise the client-side model updates.
\subsection{Model training}
Let $\ell_{\mathrm{CE}}(\cdot)$ denote the cross-entropy loss.
For a model with parameters $w$ trained on a labeled dataset $\mathcal{D}$, we define the the supervised objective function as
\begin{equation}
\label{eq:sup_erm}
F_{\mathrm{CE}}(w,\mathcal{D})
=\frac{1}{|\mathcal{D}|}\sum_{(x,y)\in\mathcal{D}}
\ell_{\mathrm{CE}}(f(w,x),y),
\end{equation}
where $f(x;w)$ denotes the model output logits.
In this paper, we adopt mini-batch stochastic optimization to minimize the supervised objective function.

The objective function in Eqn. \eqref{eq:sup_erm} will be instantiated at different stages of BiFedKD.
Specifically, each client first trains its student model on $\mathcal{D}_k$ by minimizing $F_{\mathrm{CE}}(w_k,\mathcal{D}_k)$.
This step produces a label-aware initialization that reflects the client’s own data characteristics, which makes the subsequent distillation more stable.
In each communication round, after learning on the global soft targets, the client further performs a local supervised update on $\mathcal{D}_k$, i.e., minimize $F_{\mathrm{CE}}(w_k,\mathcal{D}_k)$.
This step re-aligns the student model with the ground-truth labels, reducing possible drift caused by proxy-based distillation and maintaining personalized performance on $\mathcal{D}_k$.
On the server side, the teacher model is also warm-started by minimizing $F_{\mathrm{CE}}(w_s,\mathcal{D}_p)$ on $\mathcal{D}_p$.
This warm-up equips the teacher model with sufficient proxy-domain predictive capability, allowing it to produce reliable and informative global soft targets for top-down supervision in the following rounds.

\subsection{Client-server distillation and supervision}
At the $r$-th communication round, after training on $\mathcal{D}_k$, client $k$ computes logits on the proxy dataset $\mathcal{D}_p$ and uploads logits
$Z_k^r=[z_{k,1}^r,\,z_{k,2}^r,\,\ldots,\,z_{k,|\mathcal{D}_p|}^r]\in\mathbb{R}^{|\mathcal{D}_p|\times C}$, where $z_{k,i}^r\in\mathbb{R}^{C}$ and $C$ denotes the number of classes.
The server aggregates the uploaded logits via an element-wise average
\begin{equation}
\label{eqn_agg}
    Z^r=\frac{1}{K} \sum_{k \in \mathcal{K}} Z_k^r,
\end{equation}
where $Z^r=[z_{1}^r,\,z_{2}^r,\,\ldots,\,z_{|\mathcal{D}_p|}^r]\in\mathbb{R}^{|\mathcal{D}_p|\times C}$.
With temperature scaling factor $T>0$, the server calculates
\begin{equation}
\label{soft_label}
q_{i,c}^r = \frac{\exp(z_{i,c}^r/T)}{\sum_{c'=1}^{C} \exp(z_{i,c'}^r/T)}, \quad c=1,\dots,C.
\end{equation}
where $z_{i,c}^r$ is the logit value of the $c$-th class for $z_{i}^r$, and larger $T$ produces softer predictions.

By stacking $q_{i,c}^r$ over all proxy samples and classes, the server forms the intermediate soft target matrix $Q^r \in \mathbb{R}^{|\mathcal{D}_p| \times C}$, which is further used for teacher model distillation.
The objective function of the teacher model can be formulated as 
\begin{equation}
    H (w_s^r)= \lambda T^2 F_{\mathrm{CE}} (w_s^r)+(1-\lambda)F_{\mathrm{KL}} (w_s^r,Q^r),
\end{equation}
where
\begin{equation}
    F_{\mathrm{CE}} (w_s^r)=  \frac{1}{|\mathcal{D}_p|} \sum_{(x,y) \in \mathcal{D}_p} \ell_{\mathrm{CE}}(f(w_s^r,x),y),
\end{equation}
and
\begin{equation}
    F_{\mathrm{KL}}(w_s^r,Q^r)=  \frac{1}{|\mathcal{D}_p|} \sum_{x \in \mathcal{D}_p, q_c \in Q^r} \ell_{\mathrm{KL}}(f(w_s^r,x),q_c^r),
\end{equation}
where $\lambda \in [0,1]$ and, $w_s^r$ denotes the parameter of the teacher model at $r$-th communication round.
In the above formulations, $F_{\mathrm{CE}}(\cdot)$ quantifies the discrepancy between the teacher model’s predictive distribution and the ground-truth label distribution on $\mathcal{D}_p$,
while $F_{\mathrm{KL}}(\cdot)$ measures the discrepancy between the teacher model’s predictive distribution and the intermediate soft target $Q^r$.
By jointly minimizing $F_{\mathrm{CE}}(\cdot)$ and $F_{\mathrm{KL}}(\cdot)$, the server-side teacher model can simultaneously preserve label-discriminative capability and effectively absorb institution-specific knowledge distilled from the clients.

After distilling $Q^r$ into the teacher model, the server performs inference on $\mathcal{D}_p$ to generate the global soft target $\tilde{Q}^r$ and broadcasts it to all clients. Each client then updates its student model via distillation using $\tilde{Q}^r$. This top-down supervision encourages the student models to converge toward a globally consistent decision boundary, improving generalization under heterogeneous and non-IID ECG data distributions across institutions.

\subsection{Algorithm framework}

The comprehensive procedure of BiFedKD is presented in Algorithm 1.
As shown in lines 4-6, 
prior to the distillation stage, the server first pre-trains the teacher model on $\mathcal{D}_p$ to establish sufficient predictive capability.
Then, as indicated in lines 7-11, 
each client pre-trains its student model on $\mathcal{D}_k$, 
ensuring that the student attains a reasonable level of label-discriminative performance before participating in knowledge exchange.
Next, as shown in lines 13-16, 
all clients perform inference on $\mathcal{D}_p$ to obtain their logits and upload them to the server. 
Upon receiving the uploaded logits, the server aggregates them to form $Z^{r}$ as shown in lines 17-18, 
and converts $Z^{r}$ into the intermediate soft labels $Q^{r}$. 
In line 19, 
the server distills and updates the teacher model on $\mathcal{D}_p$, thereby integrating knowledge from all participating clients.
After the teacher model update, the server generates global soft targets $\tilde{Q}^{r}$ in line 20. Finally, as shown in lines 23-24, each client distills its student model using the received $\tilde{Q}^{r}$ on the proxy dataset, and subsequently performs a re-alignment step on the private dataset to preserve local discriminative performance.

\begin{algorithm}
  \caption{Algorithm Framework of BiFedKD}
  \begin{algorithmic}[1] 
    \STATE\textbf{Initial:} clients $\mathcal{K}$,
                            teacher model parameter $w_s^{0,0}$,
                            student model parameter $w_k^{0,0}$.
    \STATE\textbf{Input:}   Communication round $R$,
                            student model per-training epoch $E_{k}$,
                            teacher model per-training epoch $E_{s}$,
                            public proxy dataset $\mathcal{D}_p$.
                            private dataset $\mathcal{D}_k$,
    \STATE\textbf{Output:}  trained teacher model parameter $w_s^{{E_{s}},R}$,
                            trained student model parameter $ w_k^{{E_{k}},R} $.
    \FOR{$e \leftarrow 1$ to $E_{s}$}
        \STATE $w_s^{e+1,0} \leftarrow$ pre-train the teacher model.
    \ENDFOR
    \FOR{all clients in parallel}
        \FOR{$e \leftarrow 1$ to $E_{k}$}
        \STATE $w_k^{e+1,0} \leftarrow$ pre-train the student model.
    \ENDFOR
    \ENDFOR
    \FOR{$r \leftarrow 1$ to $R$}
        \FOR{all clients in parallel}
            \STATE $Z_k^r \leftarrow$ inference on $\mathcal{D}_p$,
            \STATE Send $Z_k^r$ to the server.
        \ENDFOR
        \STATE $Z^r \leftarrow $ aggregate the uploaded logits,
        \STATE $Q^r \leftarrow $ calculate the soft label,
        \STATE $w_s^{{E_{s}},r+1}\leftarrow$ distill teacher model using $Q^r$,
        \STATE $\tilde Q^r \leftarrow$ generate the soft label on $\mathcal{D}_p$,
        \STATE Send $\tilde Q^r$ to the clients.
        \FOR{all clients in parallel}
            \STATE $w_k^{{E_{k}},r+1/2}\leftarrow$ distill student model using $\tilde Q^r$,
            \STATE $w_k^{{E_{k}},r+1} \leftarrow$ re-align student model on $\mathcal{D}_k$.
        \ENDFOR
    \ENDFOR
  \end{algorithmic}
\end{algorithm}

\section{Communication Overhead and Computation Cost Analysis}
\subsection{Communication overhead analysis}
We analyze the communication overhead incurred in one communication round of the proposed BiFedKD framework.
In each round, communication takes place in two directions, 
each client uploads the logits computed on the shared proxy dataset to the server, 
and the server broadcasts the global soft targets to all clients.

Assume each scalar element in the logits or soft targets is represented by $b$ bits.
Since each proxy sample corresponds to a $C$-dimensional vector, each client uploads $|\mathcal{D}_p|C$ scalar elements and receives $|\mathcal{D}_p|C$ scalar elements in one round. 
Therefore, the per-round communication overhead of client $k$ is
\begin{equation}
\label{eq:comm_round_client}
\mathcal{C}_{\mathrm{round}}^{k} = 2b|\mathcal{D}_p|C,
\end{equation}
and the per-round communication overhead of client across all $K$ clients is
\begin{equation}
\label{eq:comm_round_total}
\mathcal{C}_{\mathrm{round}} = K\mathcal{C}_{\mathrm{round}}^{k} = 2Kb|\mathcal{D}_p|C .
\end{equation}
As shown in Eqn. \eqref{eq:comm_round_client}-\eqref{eq:comm_round_total}, the communication overhead scales linearly with the proxy set size $|\mathcal{D}_p|$, indicating that the overall communication cost of the system is highly sensitive to the choice of the proxy dataset.
In practice, the size of the proxy dataset is typically kept small, which makes the proposed BiFedKD communication-efficient and scalable to a large number of clients.

\subsection{Computation cost analysis}
We further analyze the computation overhead incurred in each communication round of BiFedKD in terms of FLOPs.
Let $\mathcal{F}_{k}^{\mathrm{train}}$ and $\mathcal{F}_{k}^{\mathrm{infer}}$ denote the per-sample training and inference FLOPs of the student model at client $k$, 
and let $\mathcal{F}_{\mathrm{s}}^{\mathrm{train}}$ and $\mathcal{F}_{\mathrm{s}}^{\mathrm{infer}}$ denote the per-sample training and inference FLOPs of the server-side teacher model.

Before communicating with the server, client $k$ conducts $E_{k}$ epochs for pre-training on its private dataset $\mathcal{D}_k$, which costs
\begin{equation}
\mathcal{G}_{k}^{\mathrm{pre}} \approx E_k|\mathcal{D}_k|\mathcal{F}_{k}^{\mathrm{train}}.
\end{equation}
In each communication round, client $k$ first infers logits on $\mathcal{D}_p$, then, distills the student model on $\mathcal{D}_p$, and finally re-aligns its model on $\mathcal{D}_k$ for one epoch. The corresponding cost is
\begin{equation}
\mathcal{G}_{k}^{\mathrm{round}}
\approx |\mathcal{D}_p|\mathcal{F}_{k}^{\mathrm{infer}}
+|\mathcal{D}_p|\mathcal{F}_{k}^{\mathrm{train}}
+|\mathcal{D}_k|\mathcal{F}_{k}^{\mathrm{train}}.
\end{equation}
On the server, the teacher model is pre-trained for $E_{s}$ epochs on the proxy dataset $\mathcal{D}_p$, and the corresponding pre-training cost is
\begin{equation}
\mathcal{G}_{\mathrm{s}}^{\mathrm{pre}} \approx E_s|\mathcal{D}_p|\mathcal{F}_{\mathrm{s}}^{\mathrm{train}}.
\end{equation}
During each communication round, the server first aggregates $K$ clients' logits and then performs one-epoch teacher distillation and one inference pass on $\mathcal{D}_p$, the corresponding computation overhead is
\begin{equation}
\mathcal{G}_{\mathrm{s}}^{\mathrm{round}}
\approx |\mathcal{D}_p|\mathcal{F}_{\mathrm{s}}^{\mathrm{train}}
+|\mathcal{D}_p|\mathcal{F}_{\mathrm{s}}^{\mathrm{infer}}
+\mathcal{O}(K|\mathcal{D}_p|C).
\end{equation}
Over $R$ rounds, the total computation cost is
\begin{equation}
\label{eq_compu}
\begin{aligned}
\mathcal{G}^{\mathrm{total}}
&\approx
\sum_{k=1}^{K}\Big(E_{k}|\mathcal{D}_k|\mathcal{F}_{k}^{\mathrm{train}}\Big)+E_{s}|\mathcal{D}_p|\mathcal{F}_{\mathrm{s}}^{\mathrm{train}}
\\
&\quad +R\Bigg[\sum_{k=1}^{K}\Big(|\mathcal{D}_p|\mathcal{F}_{k}^{\mathrm{infer}}+|\mathcal{D}_p|\mathcal{F}_{k}^{\mathrm{train}}\\
&\qquad\qquad+|\mathcal{D}_k|\mathcal{F}_{k}^{\mathrm{train}}\Big)
+|\mathcal{D}_p|\mathcal{F}_{\mathrm{s}}^{\mathrm{train}}\\
&\qquad\qquad+|\mathcal{D}_p|\mathcal{F}_{\mathrm{s}}^{\mathrm{infer}}+\mathcal{O}(K|\mathcal{D}_p|C)\Bigg].
\end{aligned}
\end{equation}
Eqn. \eqref{eq_compu} indicates that the overall cost is dominated by client-side training on $\mathcal{D}_k$ and scales linearly with $|\mathcal{D}_p|$. In addition, a heavier teacher model increases $\mathcal{F}_{\mathrm{s}}^{\mathrm{train}}$ and $\mathcal{F}_{\mathrm{s}}^{\mathrm{infer}}$, thereby enlarging the server-side overhead.

\section{Simulation Results}
\subsection{Experiment setup}
Experiments are conducted on the MIT-BIH Arrhythmia Dataset, which contains ECG recordings from 48 patients. 
We consider a federated setting with three clients and one central server.  
Each client maintains a private ECG dataset containing approximately 3,000-4,000 samples, and the corresponding class distribution and imbalance ratio (IR) is reported in Table \ref{tab:label_ratio_nsvfq_ir}, where we apply the AAMI standard \cite{1306572}. The shared proxy dataset contains 1,000 samples and is class-balanced.
Each heartbeat sample is centered at the R-peak and represented by 360 signal points.

On each client, we employ a 5-layer 1D-CNN \cite{7202837} as the student model.
On the server, a teacher model is introduced to generate the global distillation supervision. By default, we adopt a 1D-CNN-based Transformer as the teacher model, and further evaluate alternative teacher architectures in TABLE \ref{tab_server}.
The teacher pre-training epochs are set to 100, i.e., $E_{s}=100$, while the student pre-training epochs are set to 1, i.e., $E_{k}=1$.
The number of communication rounds is set to 50, i.e., $R=50$.
The distillation temperature is set to $T=3$, and the balancing coefficient is set to $\lambda=0.3$.
Adam is used to optimize both the teacher and student models. The batch size is set to 64, and the learning rates for both models are set to $2\times10^{-4}$.
In each communication round, both the uploaded logits and the broadcast global soft targets are transmitted in full-precision 32-bit floating-point format.

\subsection{Compare the performance of BiFedKD with other algorithms}
\begin{figure}[t]
  \centering
  \begin{subfigure}[b]{0.24\textwidth}
    \includegraphics[width=\textwidth]{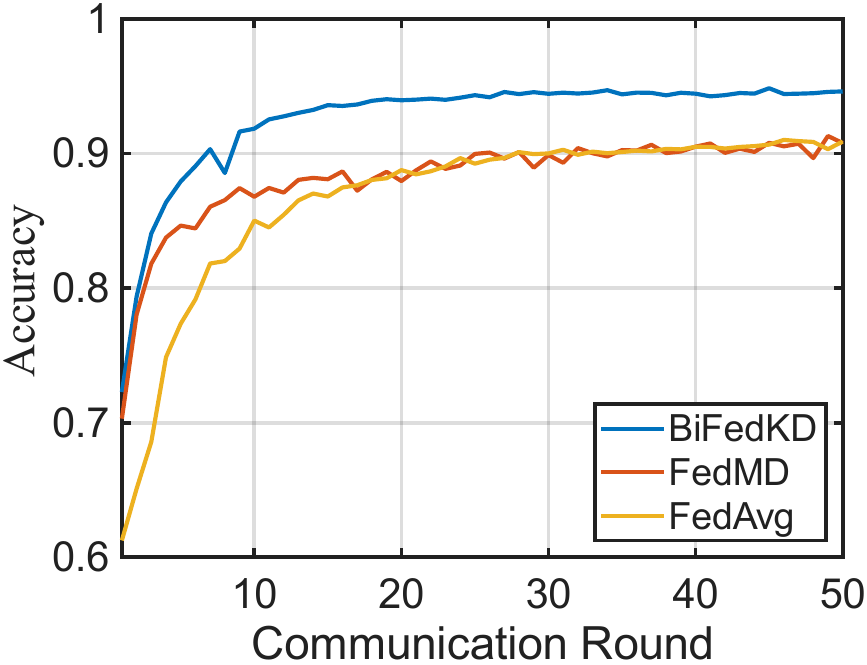}
    \caption{Accuracy}
  \end{subfigure}
  \hfill
  \begin{subfigure}[b]{0.24\textwidth}
    \includegraphics[width=\textwidth]{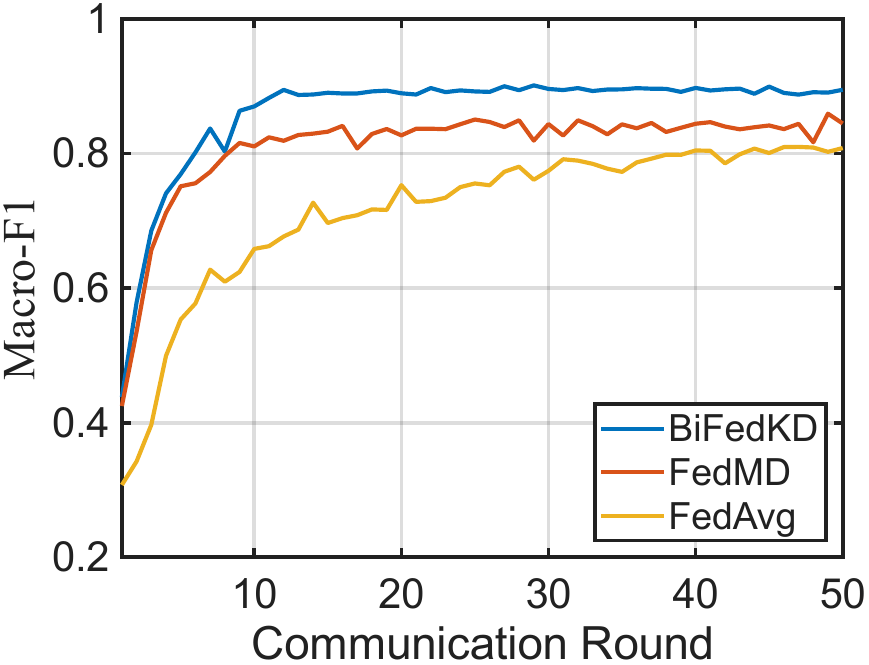}
    \caption{Macro-F1 Score}
  \end{subfigure}
  \caption{Learning curves of different algorithms.}
  \label{fig_algori_comp}
\end{figure}

As shown in Fig. \ref{fig_algori_comp}, we compare BiFedKD with FedMD and FedAvg in terms of accuracy and Macro-F1 across communication rounds. 
In Fig. \ref{fig_algori_comp}(a), 
BiFedKD converges faster and achieves consistently higher accuracy than both baselines. 
Specifically, BiFedKD reaches a stable accuracy within 25 rounds, whereas FedMD and FedAvg only stabilize after 30 rounds. 
Moreover, to attain the best accuracy achieved by FedMD and FedAvg, BiFedKD requires only about 10 rounds. 
In Fig. \ref{fig_algori_comp}(b), 
BiFedKD also maintains a clear and persistent Macro-F1 advantage over FedMD and FedAvg. 
BiFedKD and FedMD stabilize after about 15 rounds, while FedAvg converges much more slowly and only becomes stable after around 40 rounds. 
Moreover, to match the best Macro-F1 achieved by FedMD and FedAvg, BiFedKD requires only about 10 rounds.

The above advantages are primarily attributed to BiFedKD’s knowledge interaction and aggregation design. Specifically, the server-side teacher model performs distillation to aggregate and filter the logits uploaded by clients, compressing heterogeneous and potentially noisy predictions into a more consistent and stable global supervision signal, thereby mitigating the adverse impact of individual client bias or noisy uploads on the quality of global supervision. 

\begin{table}[t]
\caption{Per-client label distribution under the AAMI standard.}
\label{tab:label_ratio_nsvfq_ir}
\centering
\renewcommand{\arraystretch}{1.05}
\begin{tabular}{ccccccc}
\hline
Client & N (\%) & S (\%) & V (\%) & F (\%) & Q (\%) & IR \\
\hline
A & 31.96 & 14.46 & 43.92 & 8.19  & 1.47  & 29.82 \\
B & 48.53 & 3.82  & 36.64 & 10.59 & 0.42  & 116.80 \\
C & 29.40 & 33.45 & 32.43 & 4.72  & 0      & -- \\
\hline
\end{tabular}
\end{table}

\subsection{Analyze the communication overhead and computation cost for clients}
\begin{figure}[t]
  \centering
  \begin{subfigure}[b]{0.38\textwidth}
    \includegraphics[width=\textwidth]{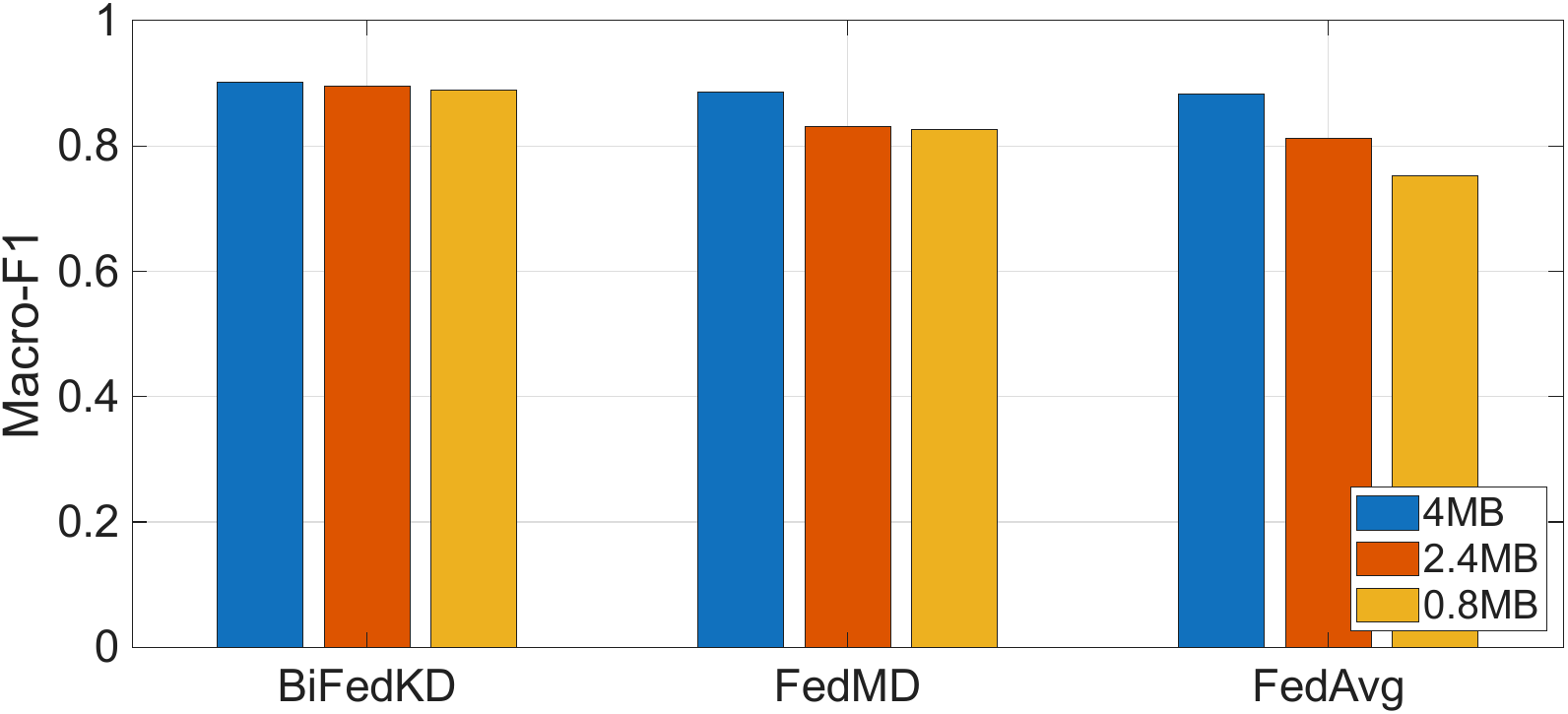}
    \caption{Macro-F1 Score under Different Communication Budgets}
  \end{subfigure}

  \begin{subfigure}[b]{0.38\textwidth}
    \includegraphics[width=\textwidth]{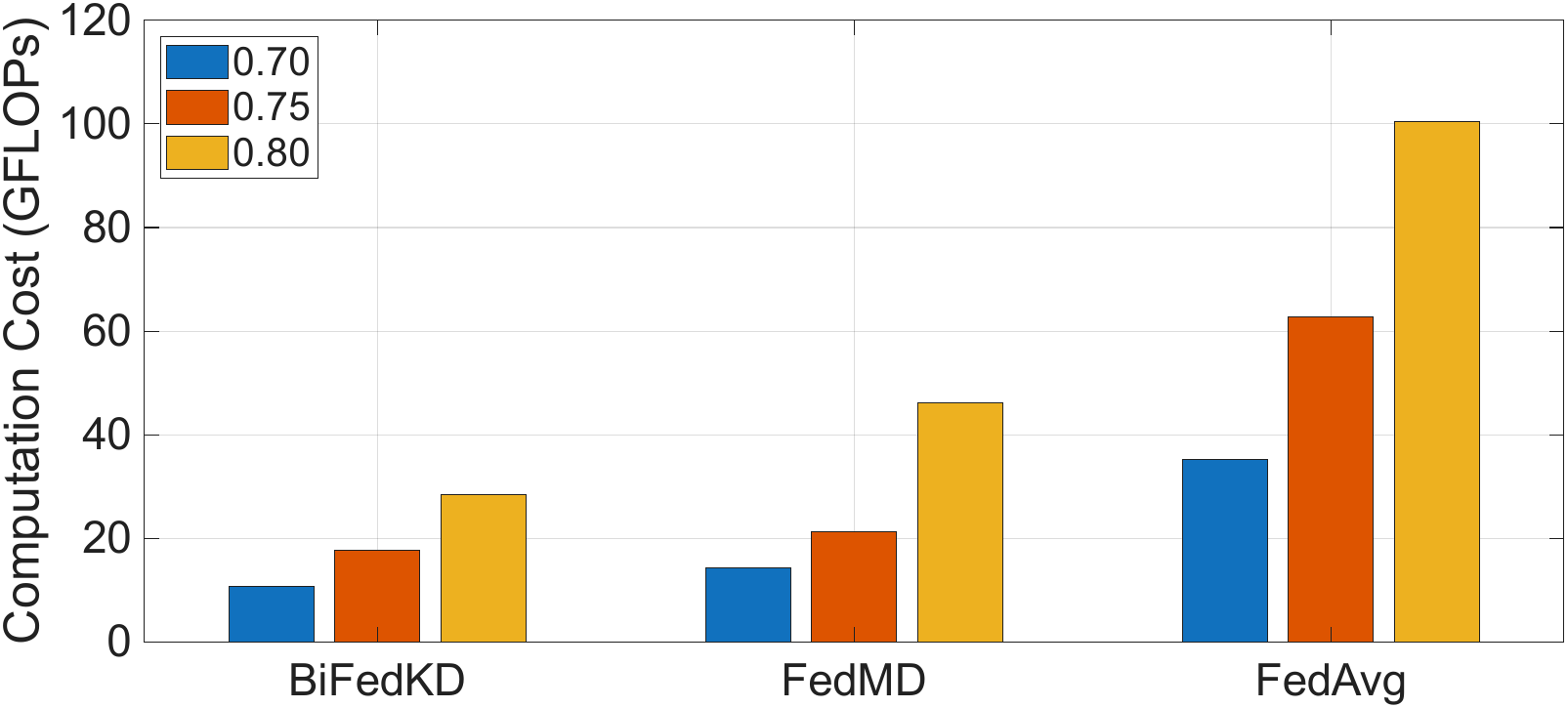}
    \caption{Computation Cost Needed to Reach Target Macro-F1 Score}
  \end{subfigure}
  \caption{(a) Communication and (b) computation efficiency comparison in terms of Macro-F1.}
  \label{fig_commu_compu}
\end{figure}

In Fig. \ref{fig_commu_compu}, we analyze and compare BiFedKD with FedMD and FedAvg under different communication budgets and target performance constraints.
Specifically, We meet the budget by adjusting communication rounds and record the relationship between Macro-F1 and the per-round per-client average communication overhead, as well as the relationship between Macro-F1 and the per-round per-client average computation overhead.
As shown in Fig. \ref{fig_commu_compu}(a),
under three communication budgets, 
BiFedKD consistently achieves the highest Macro-F1 and exhibits the smallest performance degradation as the budget decreases, 
indicating stronger robustness to communication constraints. 
Moreover, reducing the communication budget does not result in reduction of Macro-F1. 
FedMD and FedAvg both underperform BiFedKD across all communication budgets. In addition, FedAvg is more sensitive to budget reductions, exhibiting a more pronounced Macro-F1 drop under the lowest budget.
As shown in Fig. \ref{fig_commu_compu}(b),
for target Macro-F1 thresholds of 0.70, 0.75, and 0.80, 
BiFedKD requires the lowest computation cost to reach each target, 
followed by FedMD with a moderate cost. 
FedAvg incurs the highest computation, and its cost increases more rapidly as the target becomes more stringent.
Overall, these results demonstrate that BiFedKD delivers better performance under limited communication budgets and achieves the same target performance with lower computation cost, highlighting its advantage in both communication efficiency and training efficiency.

\subsection{Analyze the communication overhead and computation cost for the server}
In TABLE \ref{tab_server}, we compare teacher models with different architectures by measuring the teacher’s per-round computation cost and evaluating the resulting clients’ average accuracy and Macro-F1. 
The results show that CNN-5 offers a favorable trade-off between performance and computation overhead, while CNN Transform achieves the best accuracy and Macro-F1. CNN-7 incurs the highest computation cost yet underperforms CNN Transform, whereas CNN-3 remains competitive under tight compute budgets. 
These results indicate that the proposed bidirectional distillation enables effective cross-client knowledge transfer without requiring expensive server-side computation, thereby improving scalability and resource efficiency. 
Moreover, for practical deployments, lightweight teachers such as CNN-3 or CNN-5 can be preferred to further reduce server-side computation with only marginal degradation in clients’ accuracy and Macro-F1.

\section{Conclusion}
In this paper, we proposed a bidirectional federated knowledge distillation framework tailored for resource-constrained IoMT systems. By introducing a server-side teacher supervision mechanism, BiFedKD enhanced cross-client knowledge alignment under long-tailed and non-IID ECG data distributions. Experiments on the MIT-BIH Arrhythmia Dataset demonstrated that BiFedKD could effectively mitigate class imbalance and data heterogeneity, while incurring lower communication and computation costs.

\begin{table}[t]
\centering
\caption{Comparison of teacher models in terms of accuracy, Macro-F1 score, and computation cost.}
\label{tab_server}
\begin{tabular}{lcccc}
\toprule
                            & CNN Transform         & CNN-7             & CNN-5             & CNN-3\\
\midrule
Accuracy (\%)               & 94.75                 & 94.44             & 94.21             & 93.81  \\
Macro-F1                    & 0.8956                & 0.8828            & 0.8775            & 0.8764 \\
Cost (GFLOPs)               & 48.11                 & 202.28            & 1.04              & 0.64   \\
\bottomrule
\end{tabular}
\end{table}

\bibliographystyle{IEEEtran}
\bibliography{reference}

\end{document}